\pdfoutput=1

\documentclass[11pt]{article}

\usepackage{acl}

\usepackage{times}
\usepackage{latexsym}
\usepackage{booktabs}
\usepackage{adjustbox}
\usepackage{amsmath}
\usepackage{url}
\usepackage{geometry}
\usepackage{xcolor}
\usepackage{array}
\usepackage[skins]{tcolorbox}
\usepackage{CJKutf8}

\usepackage[T1]{fontenc}

\usepackage[utf8]{inputenc}

\usepackage{microtype}

\usepackage{inconsolata}
\usepackage{graphicx}
\newcommand*{\img}[1]{%
    \raisebox{-.01\baselineskip}{%
        \includegraphics[
        height=0.8\baselineskip,
        width=0.8\baselineskip,
        keepaspectratio,
        ]{#1}%
    }%
}
\newcommand*{\imgtext}[1]{%
    \raisebox{-.01\baselineskip}{%
        \includegraphics[
        height=0.7\baselineskip,
        width=0.7\baselineskip,
        keepaspectratio,
        ]{#1}%
    }%
}

\title{NILE~\img{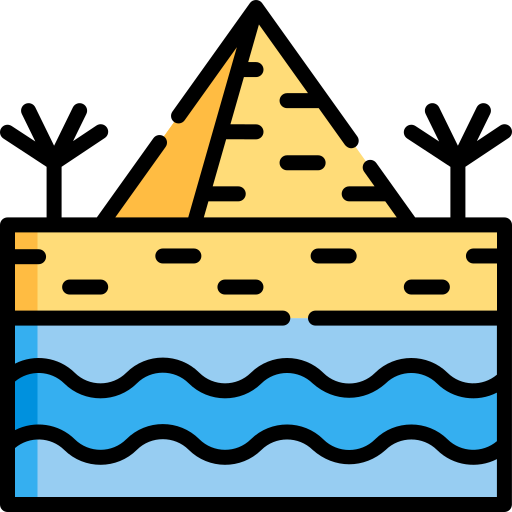}~: Internal Consistency Alignment in Large Language Models}

\author{
  Minda Hu$^{1}$, Qiyuan Zhang$^{2}$, Yufei Wang$^3$, Bowei He$^2$, Hongru Wang$^4$\\
  \bf
  Jingyan Zhou$^1$, Liangyou Li$^3$, Yasheng Wang$^3$, Chen Ma$^2$, Irwin King$^1$\\
  $^1$The Chinese University of Hong Kong
  $^2$City University of Hong Kong\hspace{0.3cm} \\
  $^3$Huawei Noah’s Ark Lab
  $^4$University of Edinburgh \\
  \texttt{\{mindahu21, king\}@cse.cuhk.edu.hk}
  }

\begin{document}
\maketitle
\begin{abstract}

Recent advances show that the world knowledge in the Instruction Fine-Tuning (IFT) dataset, which is incompatible with LLMs' internal knowledge, can greatly hurt the IFT performance. 
However, the effective integration and balancing of the internal knowledge of LLMs, acquired during pre-training, with existing IFT datasets remains a largely underexplored area of research.
To address this gap, this work introduces NILE, a novel framework to optimize the effectiveness of IFT by adjusting IFT datasets through carefully aligning the world and internal knowledge. 
NILE employs a three-stage pipeline to effectively quantify and adjust consistency with the internal knowledge of target LLMs.
Our analysis provides compelling evidence that balancing such consistency with pre-trained internal knowledge is pivotal for unleashing LLM potential, and confirms that \textsc{NILE} can systematically contribute to these substantial performance improvements. 
Experimental results demonstrate that \textsc{NILE}-aligned IFT datasets sharply boost LLM performance across multiple LLM ability evaluation datasets, achieving up to $66.6\%$ gain on Arena-Hard and $68.5\%$ on Alpaca-Eval V2. 
\end{abstract}

\section{Introduction}

Instruction Fine-Tuning (IFT), which fine-tunes Large Language Models (LLMs) on instruction-response pairs, has been proven to be an effective and crucial method to enhance the capabilities and controllability of LLMs~\citep{touvron2023llama, dubey2024llama,zhang2023instruction,chen-etal-2024-iteralign,wang-etal-2025-self-reasoning}. Most IFT approaches predominantly focus on the quantity and diversity of datasets, based on the assumption that a greater size of instruction-response pairs would lead to better performance~\citep{honovich2023unnatural, wang2023self, taori2023alpaca, chiang2023vicuna, sun2024principle}.
These approaches narrowly emphasize data quantity while overlooking IFT's core purpose: unlocking the latent capabilities of pre-trained LLMs. They do not adequately consider underlying correlations between IFT datasets and LLMs, which is crucial to the efficacy of IFT~\cite{ren-etal-2024-learning}.

\begin{figure}[!t]
    \centering
    \includegraphics[width=0.45\textwidth]{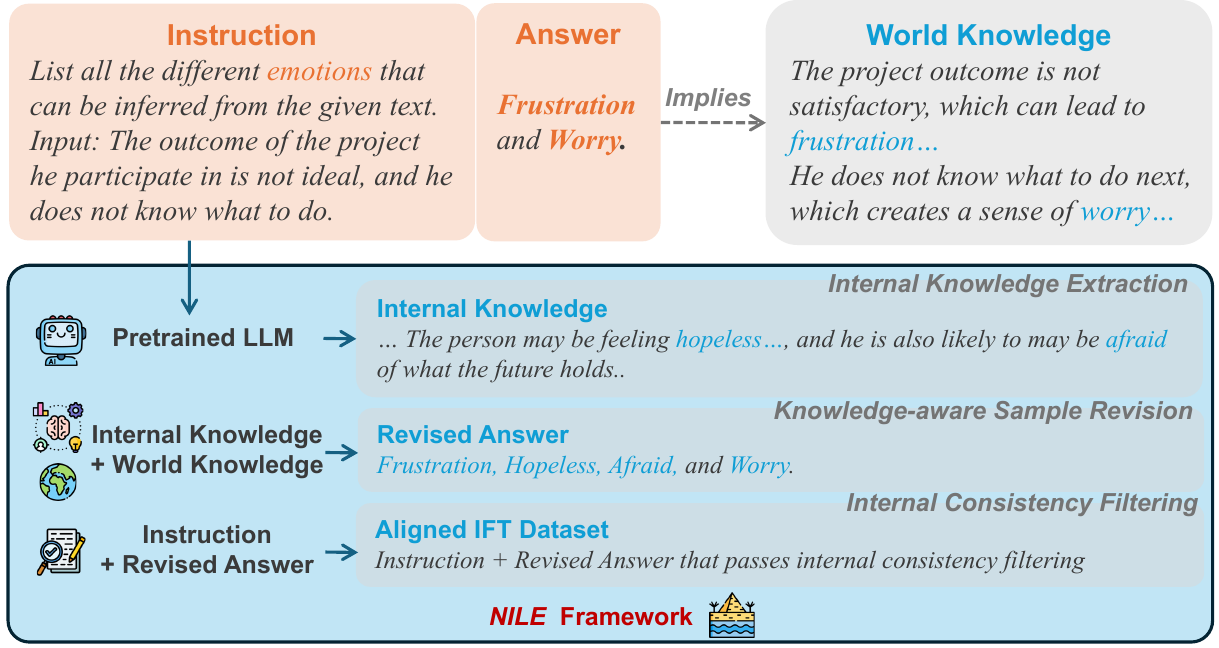}
    \caption{Demonstration of LLM internal knowledge and world knowledge from IFT datasets. 
    }
    \label{fig:word_internal_knowledge}
    \vspace{-0.5cm}
\end{figure}

A key factor influencing IFT performance is the level of \textit{internal consistency}, i.e., the consistency between the world knowledge in IFT datasets and the internal knowledge embedded within LLM parameters~\cite{ren-etal-2024-learning}.
When trained on totally unfamiliar data, i.e., data with low internal consistency, LLMs may only capture superficial correlations in instruction-response pairs, such as text styles, and tend to make ``blind guesses'' when faced with new queries \cite{kang2024unfamiliar}.
Nonetheless, \citet{ren-etal-2024-learning} shows that merely maximizing internal consistency does not necessarily lead to optimal IFT performance.
These works suggest that examining and curating the internal consistency of IFT datasets for the target pre-trained LLM is a promising direction for effective training.
However, how to revise and balance the internal consistency level remains under-explored.

In this work, we propose a novel framework  \textsc{\textbf{NILE}\imgtext{figures/nile.png} (I\textbf{N}ternal Cons\textbf{I}stency a\textbf{L}ignm\textbf{E}nt)}.
NILE bridges the aforementioned research gap by flexibly improving existing IFT datasets in terms of internal consistency for the target pretrained LLM.
Specifically, NILE addresses the problem through the following three steps: \textit{1) Internal Knowledge Extraction.} As a prerequisite, accurately extracting internal knowledge is crucial. We adopt in-context learning techniques with high-quality customized examples. \textit{2) Knowledge-aware Sample Revision.} To fully utilize existing data, we designed a revision step to improve the existing data with LLMs' internal knowledge, resulting in a data sample with higher consistency. \textit{3) Internal Consistency Filtering.} Lastly, we developed a novel metric to measure the consistency level between the data sample and the LLM.
By doing so, we can flexibly adjust the level of internal consistency of existing IFT data with any target pre-trained LLM to achieve optimized IFT performance.

It is important to highlight that our method does not rely on any additional forms of supervision (i.e., human experts). To conclude, our contributions can be summarized as follows:


\begin{itemize}
    \item We propose NILE, a novel framework to adjust and select better IFT datasets considering the consistency between internal parameter knowledge in LLMs and world knowledge in IFT datasets, as shown in Figure~\ref{fig:word_internal_knowledge}. To the best of our knowledge, we are among the first to leverage the concept of internal consistency for IFT data selection and generation.\footnote{Corresponding NILE-revised IFT datasets can be found in \url{https://huggingface.co/datasets/mindahu/NILE-IFT-Dataset}.} 
    \item Through comprehensive ablation studies and empirical analysis, we demonstrate that balancing consistency between IFT datasets and LLMs' internal knowledge is crucial for unlocking model capabilities. Our results provide strong evidence that each component of NILE contributes to performance gains.
    \item  Our extensive experiments across multiple benchmarks show that NILE-optimized datasets enable substantial improvements in LLM performance, achieving up to 66.6\% gains on Arena-Hard and 68.5\% on Alpaca-Eval V2. These results demonstrate that NILE's balanced integration of world and internal knowledge enhances LLMs' ability to generalize to novel tasks and domains.
\end{itemize}

\section{Related Works}

\subsection{Data Synthesis in Instruction Tuning }

Earlier research on instruction tuning has primarily focused on developing large, high-quality datasets curated by human experts \citep{wei2022finetunedlanguagemodelszeroshot, wang-etal-2022-super}. However, this process is often time-consuming and labor-intensive. Thus, several studies have explored the use of more advanced models or self-critique prompting methods \citep{wang-etal-2023-cue, wang-etal-2024-enhancing, zhang2024selftuninginstructingllmseffectively,pi2024image,pi2024personalized} to generate instruction-tuning datasets automatically. For example, Self-Instruct \citep{wang2023self} leverages GPT-3 to expand asks to many diverse domains in an in-context learning manner while several recent studies directly use the latest SOTA model to generate the response or reflect on current samples \cite{mukherjee2023orca}, such as WizardLM \cite{xu2023wizardlm} and Reflection-tuning \cite{DBLP:conf/acl/LiCCHGZ24}. In addition to focusing on the quality side, another area of work aims to create more diverse and larger instruction-tuning datasets. For example, UltraChat \cite{ding-etal-2023-enhancing} defines specific scopes and systematically generates a wide range of instructions within each area. In contrast, Magpie \citep{xu2024magpiealignmentdatasynthesis} only feeds the left-side templates up to the position reserved for user messages as input to generate more diverse user queries.

For complex reasoning tasks such as coding and mathematics, many efforts have been made to integrate human priors into data synthesis~\cite{zhou2025survey}. KPDDS~\cite{huang2025key} leverages key points and exemplar practices to synthesize mathematical reasoning-focused IFT datasets. Additionally, Case2Code~\cite{shao2024case2code} introduces observations of input-output examples and program behaviors to infer underlying code implementations.

\subsection{Data Selection in Instruction Tuning }

Data selection (or revision) has been widely studied in large language model instruction tuning, considering the importance of data quality in model training~\citep{li2024quantity, caoinstruction, li2024nuggets, zhou2024lima, liu2024coachlm, DBLP:conf/acl/LiCCHGZ24}. Most previous studies fall into two categories: 1) relying on more powerful models or human experts to select better data \citep{zhou2024lima, liu2024coachlm}; 2) calculating the perplexity gains considering generated samples and original samples \citep{DBLP:conf/acl/LiCCHGZ24}. While both methods improve downstream performance, they face significant limitations, such as the high cost of human labeling. More importantly, such studies~\citep{chenself, liself, sun2024principle} can not provide fundamental explanations regarding the key factors that define better instruction-tuning datasets. In contrast to these approaches, our work aligns the internal knowledge of LLMs with external world knowledge derived from IFT datasets, resulting in improved datasets that offer better explainability and transparency.

\begin{figure*}[!ht]
    \centering
    \includegraphics[width=0.88\textwidth]{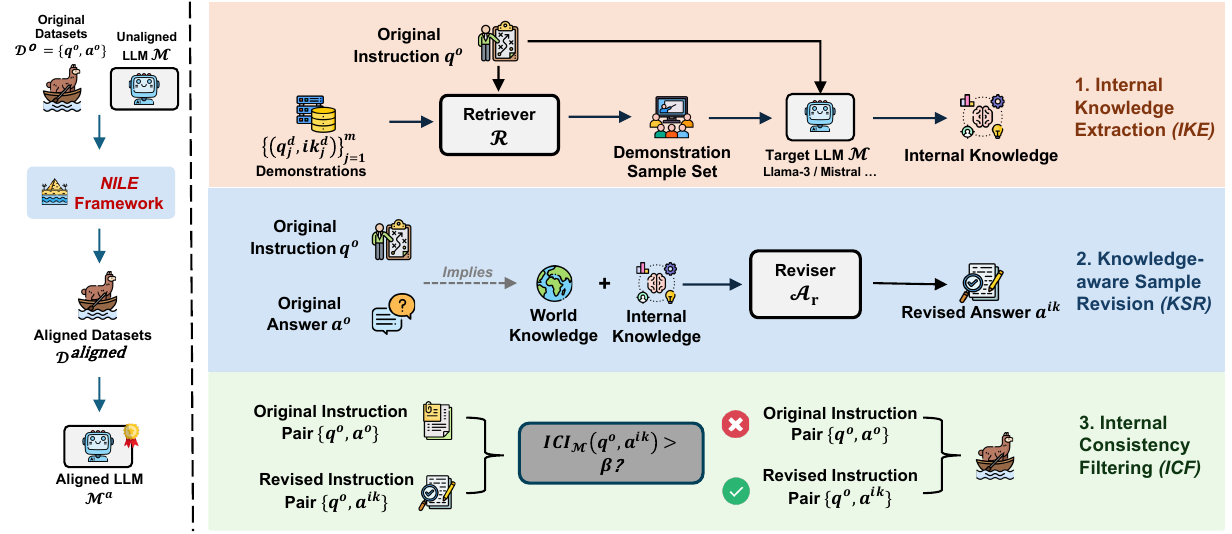}
    \caption{Overview of our \textsc{NILE} framework. \textsc{NILE} consists of three main steps: \textit{Internal Knowledge Extraction (IKE)}, \textit{Knowledge-aware Sample Revision (KSR)}, and \textit{Internal Consistency Filtering (ICF)}.}
    \label{fig:nile_overview}
    \vspace{-0.3cm}
\end{figure*}
\section{Method}
\label{sec:method}
Figure ~\ref{fig:nile_overview} demonstrates our framework \textsc{NILE} for increasing knowledge affinity between LLMs' internal knowledge and instruction-tuning datasets. It can be divided into three parts: (1) \textit{\textbf{I}nternal \textbf{K}nowledge \textbf{E}xtraction (\textbf{IKE})}, (2) \textit{\textbf{K}nowledge-aware \textbf{S}ample \textbf{R}evision (\textbf{KSR})}, and (3) \textit{\textbf{I}nternal \textbf{C}onsistency \textbf{F}iltering (\textbf{ICF})}. The core contribution of our framework lies in our deliberate focus on internal consistency, which enables the process to function effectively. IKE accesses the memory of pretrained LLMs
to sample their internal knowledge. KSR revises existing dataset samples by automatically infusing the sampled internal knowledge. ICF introduces a novel internal consistency measurement to filter out low-quality revisions from the second phase. 
In the following subsections, we introduce the above three components in detail. Implementation details of IKE, KSR, and ICF are listed in Appendix~\ref{sec:imple_nile}.
\subsection{Internal Knowledge Extraction}
This stage aims to effectively sample the internal knowledge from the target pre-trained LLM $\mathcal{M}$ for instructions in the original IFT dataset $\mathcal{D}^o = \{(q^{o}_i,\ a^{o}_i)\}_{i=1}^n$, where $q^o_i$ is the concatenated query sequence of $\mathrm{\textbf{instruction}}^{o}_i$ and $\mathrm{\textbf{input}}^{o}_i$, and $a^o_i$ is the answer.
Formally, we aim to sample the internal knowledge $ik_{i}$ corresponding to $q^o_{i}$ from $\mathcal{M}$ through in-context learning.
Instead of using a fixed set of examples, we use following three-step strategy to provide the most relevant examples to better exert the internal knowledge from $\mathcal{M}$.
\begin{enumerate}
\item \textbf{Demonstration set construction:} we first randomly sample a subset of queries $\{q_j^{d}\}_{j=1}^m$ from an IFT dataset.
Then, as shown in Table~\ref{tab:demo_gen_prompt},  a strong LLM (GPT-4 utilized in the experiments) is prompted to generate the corresponding knowledge snippet $ik^{d}_j$ for each $q_j^d$, resulting 
in a demonstration database index $\mathcal{F}^{demo} = \{(q^{d}_j, ik^{d}_j)\}_{j=1}^m$.
Details are provided in Appendix~\ref{sec:ike_sample_demon}.
\item \textbf{Example selection:} for each query $q^{o}_{i} \in \mathcal{D}^o$, we select $k$ few-shot examples $f^{\mathcal{R}}(q_i^o) = \{(q^{d}_{i_t}, ik^{d}_{i_t})\}_{t=1}^k$ from $\mathcal{F}^{demo}$, where $({i_t})_{t=1}^k$ denotes the indices of top-$k$ example pairs ranked by the query semantic similarity between $\{q_j^{d}\}_{j=1}^m$ and $q^{o}_{i}$ from 
retriever $\mathcal{R}$.
$\mathcal{R}$ is implemented by information retrieval algorithms such as BM25.
\item \textbf{Internal knowledge generation:} we formulate the prompt shown in Table~\ref{tab:ik_extract_demo} to $\mathcal{M}$ with few-shot examples $f^{\mathcal{R}}(q^o_i)$ and the original instruction $q^o_i$.
By this means, it can effectively exert internal knowledge $ik_{i}$ from the target LLM $\mathcal{M}$.
\end{enumerate}


\begin{table}[!ht]
\small
    \centering
    \begin{tcolorbox}[colframe=black, colback=gray!10!white, coltitle=black, boxrule=0.5mm]
    Generate a list of related knowledge about the following Instruction and input up to 500 words. Do not directly output the answer, but focus on the related knowledge required for answering the Input.\\
    \\
    Instruction: 
    "$\mathbf{\{\mathrm{\textbf{instruction}}^{d}_j\}}$"\\
    Input: 
    "$\mathbf{\{\mathrm{\textbf{input}}^{d}_j\}}$"
    \end{tcolorbox}
    \caption{Prompt for demonstration set construction.}
    \label{tab:demo_gen_prompt}
    \vspace{-0.3cm}
\end{table}

\begin{table}[!ht]
\small
    \centering
    \begin{tcolorbox}[colframe=black, colback=gray!10!white, coltitle=black, boxrule=0.5mm]
    \textbf{$\mathbf{\{(q^{d}_{i_t}, ik^{d}_{i_t})\}_{t=1}^k}$} 
    \\
    \\
    Instruction:\\
    $\mathbf{\{\mathrm{\textbf{instruction}}^{o}_i, \mathrm{\textbf{input}}^{o}_i\}}$\\
    \\
    Related Knowledge:
    \end{tcolorbox}
    \caption{Prompt for knowledge extraction. Sample few-shot demonstration prompt is listed in~\ref{sec:ike_sample_demon}.}
    \label{tab:ik_extract_demo}
    \vspace{-0.3cm}
\end{table}
By following this approach, we can effectively extract the internal knowledge of unaligned LLMs relevant to the original instructions, leveraging the power of few-shot demonstration learning.

\subsection{Knowledge-aware Sample Revision}
After obtaining a relatively accurate sampling $ik_i$ of the target LLM's internal knowledge (analyzed in Section~\ref{sec:ablation_st}), for each original instruction $q^o_i$, we design a prompt for the revisor LLM agent $\mathcal{A}_r$ to infuse $ik_i$ into the current instruction and get the revised answer $a^{ik}_i$. The prompt for KSR is displayed in Table~\ref{tab:sample_rev_prompt}.
\begin{table}[!ht]
\small
    \centering
    \begin{tcolorbox}[colframe=black, colback=gray!10!white, coltitle=black, boxrule=0.5mm]
    Provide a better response based on "$\mathbf{\{a^o_i\}}$" to comply with given instruction, input, and related knowledge.\\\\
    Instruction: $\mathbf{\{\mathrm{\textbf{instruction}}^{o}_i\}}$\\
    Input:$\mathbf{\{ \mathrm{\textbf{input}}^{o}_i\}}$\\
    Related Knowledge: $\mathbf{\{ik_{i}\}}$\\\\
    Please directly output the improved response.
    \end{tcolorbox}
    \caption{Prompt for Knowledge-aware Sample Revision.}
    \vspace{-0.3cm}
    \label{tab:sample_rev_prompt}
\end{table}

This step aims to enhance affinity between the target model $\mathcal{M}$'s internal knowledge $ik_{i}$ and the original answer $a^{o}_i$ from $\mathcal{D}^o$ with world knowledge, resulting an improved answer $a^{ik}_i$.
\subsection{Internal Consistency Filtering}
In this stage, we evaluate the effectiveness of KSR by comparing the quality of the revised answer $a^{ik}_i$ with the original answer $a^o_i$. Drawing inspiration from IFD and PMI~\cite{li2023quantity}, we introduce a novel metric called \textsc{\textbf{I}nternal \textbf{C}onsistency \textbf{I}ndex (\textbf{ICI})}
to quantify how well one answer promotes knowledge associations in the pretrained LLM $\mathcal{M}$.

During the instruction alignment process, the loss of a sample pair $(q, a)$ is computed using the sequence probability of $a$ conditioned on $q$:
\begin{equation}
\begin{aligned}
P_{\mathcal{M}}&(a \mid q)
 =\\&\frac{1}{N}\sum_{i=1}^N \log P_{\mathcal{M}}\left(w_i^a \mid q, w_1^a, w_2^a, \ldots, w_{i-1}^a\right),
\end{aligned}
\end{equation}
where $w_i$ is the tokens in $a$ and $N$ is the sequence length of $a$. This probability measures the familiarity of $\mathcal{M}$ with answer $a$ given the context $q$. It can also reflect the strength of the encoded association between $a$ and $q$ in the LLM's representations, which is empirically supported by \citeauthor{kang2024unfamiliar}. Building upon this idea, we formulate ICI as follows:
\begin{equation}
\begin{aligned}
\operatorname{ICI}_{\mathcal{M}}(q, a^{ik})=  \frac{P_{\mathcal{M}}(a^{ik} \mid q, ik)}{P_{\mathcal{M}}(a^{ik} \mid q)},
\end{aligned}
\end{equation}
where $P_{\mathcal{M}}(a^{ik} \mid q)$ measures the associations between revised responses $a^{ik}$ and instructions $q$ alone, while $P_{\mathcal{M}}(a^{ik} \mid q, ik)$ captures the overall association strength between $a^{ik}$ and the combination of $q$ and its corresponding extracted internal knowledge $ik$. To isolate the influence of $ik$ on the revised answer $a^{ik}$, we minimize the influence of $q$ in the ICI formulation by dividing $P_{\mathcal{M}}(a^{ik} \mid q, ik)$ with $P_{\mathcal{M}}(a^{ik} \mid q)$.

For samples with higher ICI, the model more effectively integrates and leverages the explicitly provided internal knowledge when generating the revised answer, suggesting a stronger alignment between the revised answer and the model's internal knowledge.
Conversely, for samples with lower ICI, providing internal knowledge may not benefit or could even hinder the generation of the revised answer, indicating that the revised answer does not have a strong association with what the model has learned internally, as suggested by \citet{ren-etal-2024-learning}. Therefore, we employ a filtering 
mechanism ICF to filter out these redundant low ICI samples to an aligned dataset $\mathcal{D}^{aligned}$ for fine-tuning an aligned LLM $\mathcal{M}^a$ from $\mathcal{M}$. To control dataset size in the experiment and ensure stable improvement, we revert to the original samples $(q, a^o)$ when the ICI values of $(q, a^{ik})$ are lower than the threshold $\beta$:
\begin{equation}\label{eq:icf}
    \begin{aligned}
        & \mathcal{D}^{aligned} = \{q^o_{1...n}, a^{aligned}_{1...n}\}, \\
        & \mathrm{where}\ a^{aligned}_{i} = \left\{\begin{array}{ll}
         a_i^{ik}\mbox{, if } \operatorname{ICI}_{\mathcal{M}}(q^o_i, a^{ik}_i) > \beta\\
         a_i^o\mbox{, otherwise}
         \end{array}\right.
    \end{aligned}
\end{equation}
Here we use $\beta$ to control the degree of internal consistency in ICF.

\section{Experiments}
For the main experiment, we use open source models like \textsc{Mistral-7b-v0.3}~\cite{jiang2023mistral} and \textsc{Meta-Llama-3.1-8b}~\cite{dubey2024llama} on two public datasets Alpaca~\cite{taori2023alpaca} and OpenOrca~\cite{mukherjee2023orca} to examine NILE framework's robustness extensively. In addition, we conduct an ablation study to evaluate the efficacy of our design choices in the pipeline. More experiment details, ablation study, inference overhead, and case studies can be found in ~\ref{sec:imple_nile}.
\subsection{IFT Datasets}
\paragraph{Alpaca}
The Alpaca dataset contains 52,000 instruction-following data generated using the techniques in the Self-Instruct~\cite{wang-etal-2023-self-instruct}. It starts with a limited (e.g., $175$ in our study) seed set of manually written tasks that are used to guide the overall generation. Then language models are utilized and prompted to augment these instructions and create corresponding instruction-answer instances. In our experiments, we use all the samples in a newer version of Alpaca\footnote{\url{https://huggingface.co/datasets/vicgalle/alpaca-gpt4}} dataset,  which includes instruction-following instances generated using GPT-4~\cite{peng2023instruction}.

\paragraph{Orca}
OpenOrca is a large-scale dataset built upon the Flan 2022 Collection~\cite{mukherjee2023orca,longpre2023flan}. In the Orca dataset, query-response pairs are augmented with detailed responses from GPT-4 that explain the reasoning process of the teacher as it generates the response. In contrast with vanilla instruction tuning methods like Alpaca providing little opportunity for mimicking the ``thought'' process, this dataset provides additional signals for learning to elicit such explanations. For experiments, we use the officially released dataset\footnote{\url{https://huggingface.co/datasets/Open-Orca/1million-gpt-4}}, and randomly select $50,000$ sample pairs from a pool of 1 million samples.
\subsection{Evaluation}
We briefly introduce evaluation methods used in our experiments as follows.
\paragraph{Arena-Hard (A.-H.)}
Arena-Hard-Auto\footnote{\url{https://github.com/lmarena/arena-hard-auto}} is a popular open-ended evaluation tool for instruction-tuned LLMs~\cite{li2024crowdsourced}. It contains 500 challenging user queries. GPT-4-Turbo is prompted as a judge to compare the models' responses against a baseline model. Notably, Arena-Hard keeps a high correlation and separability to Chatbot Arena~\cite{chiang2024chatbot}.

\paragraph{Alpaca-Eval V2 (A.-E. V2)} 
Alpaca-Eval V2\footnote{\url{https://github.com/tatsu-lab/alpaca_eval}} is an automatic evaluation system for instruction-following language models~\cite{dubois2024length}. It builds upon the original AlpacaEval system, which benchmarked against OpenAI's Davinci-003. AlpacaEval V2 instead uses GPT-4-Turbo, signaling the new state-of-the-art model since the original system's creation. 

A key innovation in Alpaca-Eval V2 is the introduction of \textbf{L}ength-\textbf{C}ontrolled \textbf{W}in \textbf{R}ates~(\textbf{LCWR}). It increases the correlation with ChatBot Arena to 0.98, significantly decreasing length gameability in comparison with the original \textbf{W}in \textbf{R}ate~(\textbf{WR}). In presenting experimental results, we display reports both metrics in the format: \textbf{LCWR / WR}. This provides a more comprehensive picture of model performance, with \textbf{LCWR} serving as the primary metric while still allowing comparison to the original \textbf{WR} scores.

\paragraph{MTBench (MTB.)}
MT-Bench comprises $80$ multi-turn questions spanning eight distinct knowledge domains. The models are required to respond to an initial question and subsequently provide a second response to a follow-up question. GPT-4 assesses each model’s responses on a scale from 1 to 10, and the overall score is determined by the mean over the two turns across all questions. We evaluate using the Fastchat implementation\footnote{\url{https://github.com/lm-sys/FastChat/blob/main/fastchat/llm_judge}}.

\paragraph{BBH}
Big Bench Hard\footnote{\url{https://github.com/EleutherAI/lm-evaluation-harness/tree/main/lm_eval/tasks/bbh}}~(BBH) is a suite of $23$ challenging BIG-Bench tasks~\cite{suzgun-etal-2023-challenging,srivastava2022beyond}. These tasks are chosen because prior language models showed performance below the average human-raters. Since many tasks in BBH require multi-step reasoning, CoT prompting is added to better depict the LLMs' capacities on these complex tasks that are challenging even for humans. 

\subsection{Implementation details}
For our experiments, we fine-tune the pretrained but unaligned models, \textsc{Mistral-7b-v0.3} and \textsc{Meta-Llama-3.1-8b}. For selecting retriever $\mathcal{R}$ in IKE, we find that BM25 is more effective than a strong neural retriever such as contriver~\cite{lei-etal-2023-unsupervised} in retrieving higher-quality demonstrations, which is evaluated and validated in Appendix~\ref{sec:bm25_vs_contriver}. To maintain a better state of internal consistency, we set $\beta$ in Eq.~\ref{eq:icf} to the 1st percentile of the ICI distribution for Alpaca and to the 2nd
percentile for OpenOrca to rule out a small amount of low ICI samples. Based on our manual random screening of 100 sample points respectively in Alpaca and OpenOrca datasets, we found the selected values in ICF to be a reasonable balance - lower thresholds retain too many misaligned knowledge samples that could directly impair performance, while higher thresholds discard too many consistent samples.
\subsection{Baselines}
\paragraph{Vanilla} Vanilla setting refers to using the original, unmodified IFT datasets for fine-tuning LLMs such as \textsc{Mistral} and \textsc{Llama-3}. This serves as a baseline to compare the effectiveness of dataset revision techniques.
\paragraph{SR} Sample Revision~(SR) marks the baseline for revising the instruction-answer pairs without leveraging any internal knowledge from the target LLM $\mathcal{M}$. This lets SR solely infuse knowledge from the revisor agent $\mathcal{A}_r$ into IFT datasets. Details of SR can be found in~\ref{sec:sr}.
\paragraph{NILE} NILE represents our complete proposed method. In the experiments, Alpaca and Orca datasets undergo a step-by-step revision process through the pipeline of IKE, KSR, and ICF introduced in Section~\ref{sec:method}.

To maintain consistency and a fair comparison with the Vanilla setting, the implementation of NILE and SR baseline rewrites only the responses $a^o$, leaving the rest of the dataset unchanged.

\begin{table*}[!ht]
    \centering
    \small
    \setlength{\tabcolsep}{3.2mm}{
    \begin{tabular}{lcccc}
    \toprule
         Method  
         &  Arena-Hard~$\uparrow$ & Alpaca-Eval V2~$\uparrow$ & MTBench~$\uparrow$ & BBH~$\uparrow$ \\
         \midrule
         \multicolumn{5}{c}{\textsc{Mistral-7b-v0.3}}\\
         \midrule
         \textsc{Alpaca vanilla}  & 3.00 & \underline{11.73} / \underline{7.39} & \underline{6.37} & 34.46 \\
         \textsc{Alpaca + SR}  & \underline{4.20} & 11.50 / 6.52 & 6.28 &  \underline{38.40} \\
         \textsc{Alpaca + NILE} & \textbf{6.20} & \textbf{15.39} / \textbf{9.70}  & \textbf{6.56} & \textbf{38.52} \\
         \midrule
         \textsc{Orca vanilla}  & 5.30 & 12.84 / 9.54 & 5.34 & \underline{46.37} \\
         \textsc{Orca + SR}  & \underline{5.70} & \underline{18.19} / \underline{15.24} & \underline{6.13} &  46.01 \\
         \textsc{Orca + NILE} & \textbf{6.70} & \textbf{21.63} / \textbf{17.25}  & \textbf{6.73} & \textbf{51.01} \\
         \midrule
         \multicolumn{5}{c}{\textsc{Meta-Llama-3.1-8B}}\\
         \midrule
         \textsc{Alpaca vanilla }  & 2.10 & 7.58 / 5.53 & 6.31 & 58.64 \\
         \textsc{Alpaca + SR}  & \underline{3.30} & \underline{9.08} / \underline{6.84} & \underline{6.39} &  \underline{59.91} \\
         \textsc{Alpaca + NILE}  & \textbf{4.80} & \textbf{10.69} / 
         \textbf{10.43} & \textbf{6.90} & \textbf{61.40} \\
         \midrule 
         \textsc{Orca vanilla}  & 3.60 & 10.84 / 7.52 & 7.01 & 63.02 \\
         \textsc{Orca + SR} & \underline{4.20} & \underline{12.36} / \underline{10.46} & \underline{7.18} & \underline{63.77} \\
         \textsc{Orca + NILE}  & \textbf{6.00} & \textbf{13.70} / \textbf{12.11} & \textbf{7.48} & \textbf{64.05} \\
        \bottomrule

    \end{tabular}
    }
    \caption{Main experiment results on Alpaca and OpenOrca datasets. The highest values are \textbf{bolded}, and the second highest is \underline{underlined}. 
    }
    \vspace{-0.4cm}
    \label{tab:main_exp}
\end{table*}
\subsection{Results on Orca Dataset}
Table~\ref{tab:main_exp} shows the performance of our \textsc{NILE} framework and all baselines on model \textsc{Mistral-7b-v0.3} and \textsc{Meta-Llama-3.1-8b} in OpenOrca dataset. As we can see, Orca dataset brings unbalanced improvements on different LLMs, with \textsc{LLama-3} having less improvements on \textbf{Arena-Hard} and \textbf{Alpaca-Eval V2 LCWR} and more on \textbf{MTBench} and \textbf{BBH} than \textsc{Mistral}, which reflects different underlying characteristics and potentially distinct internal knowledge in these two models. 

Compared with \textsc{Orca vanilla}, \textsc{Orca + NILE} brings substantial improvements on all benchmarks in both LLMs. It increases \textbf{Arena-Hard} score by 1.4 points~(26.4\% relative improvement) in \textsc{Mistral} and 2.4 points~(66.6\%) in \textsc{Llama-3}. NILE also significantly enhances \textbf{Alpaca-Eval V2 LCWR} from 12.73 to 21.63 in \textsc{Mistral} and 10.84 to 13.70 in \textsc{Llama-3}, achieving 68.5\% and 26.4\% relative improvements respectively. 

In addition, it is noteworthy that \textbf{NILE} also brings considerable boosts on \textbf{BBH} benchmark by 4.64 in \textsc{Mistral} and by 1.05 in \textsc{Llama-3}. BBH tasks mainly focus on tasks requiring complex reasoning and expert knowledge, and performance lift of \textsc{Orca + NILE} compared to \textsc{Orca vanilla} indicates the fact that alignment dataset revised by NILE encroaches fewer LLMs' innate capability of multi-step complex reasoning since instructions in OpenOrca dataset itself is barely involved with multi-step complex reasoning, and yet \textsc{Orca + NILE} helps unleashing the reasoning ability of the LLMs, as shown in the result of \textbf{BBH}. The universal improvements in these four well-tested benchmarks provide strong support for NILE's effectiveness in improving LLMs' general capacity.

Compared to \textsc{Orca + NILE}, \textsc{Orca + SR} infuses only the internal knowledge of the GPT-4 revisor model without utilizing extracted knowledge from \textsc{Mistral} and \textsc{Llama-3} or the ICF phase. The experiment involving \textsc{Orca + SR} is designed to investigate the contribution that introducing LLMs' own internal knowledge makes in the NILE framework. \textsc{Orca + NILE} largely surpasses \textsc{Orca + SR} by 3.4 and 5.0 points on \textbf{Alpaca-Eval V2 LCWR} and \textbf{BBH} in \textsc{Mistral} model, 1.3 and 1.8 points on \textbf{Alpaca-Eval V2 LCWR} and \textbf{Arena-Hard} in \textsc{Llama-3}. This indicates that internal knowledge extracted from LLMs is crucial for bringing more performance uplift in LLM's general capability.
\subsection{Results on Alpaca}
Compared with Orca dataset, LLMs finetuned with Alpaca dataset are generally weaker than ones with Orca, which highlights the sheer quality differences between the two datasets. Despite these differences, \textsc{Alpaca + NILE} still brings significant improvements over \textsc{Alpaca vanilla} in all metrics, coming close to or even surpassing \textsc{Orca vanilla} in most of the benchmarks except \textbf{BBH}. It achieves a performance uplift by 3.7 and 4.1 points on \textbf{Alpaca-Eval V2 LCWR} and \textbf{BBH} in \textsc{Mistral}. Moreover, \textsc{Alpaca + NILE} raises \textbf{Alpaca-Eval V2 LCWR} and \textbf{Arena-Hard} by 3.1 and 2.7 in \textsc{Llama-3}. 

Measured against \textsc{Alpaca + SR}, \textsc{Alpaca + NILE} still maintains major advantages. It enhances \textbf{Arena-Hard} and \textbf{Alpaca-Eval V2} by 2.0 and 3.9 in \textsc{Mistral} model, 1.5 and 1.6 in \textsc{Llama-3}. These results further illustrate the necessity of extracting internal knowledge in NILE.

\subsection{Experiment Results on More LLMs}
We conduct additional experiments of NILE on pretrained models of varying sizes, such as \textsc{Meta-Llama-3.2-3B}\footnote{\url{https://huggingface.co/meta-llama/Llama-3.2-3B}}, as well as on different model families, including \textsc{Qwen2.5-7B}\footnote{\url{https://huggingface.co/Qwen/Qwen2.5-7B}} and \textsc{Qwen2.5-14B}\footnote{\url{hhttps://huggingface.co/Qwen/Qwen2.5-14B}}, using the 
\textbf{MTBench} and \textbf{Alpaca-Eval V2} benchmarks. The consistent and significant relative improvements (up to 85.7\% on \textbf{Alpaca-Eval V2 LCWR}) observed in Table~\ref{tab:more_exp_on_llm} demonstrate that NILE consistently delivers meaningful improvements across diverse LLM configurations.
\begin{table}[!ht]
    \centering
    \small
    \setlength{\tabcolsep}{1.6mm}{
    \begin{tabular}{lcc}
    \toprule
         Method  
        &  MTBench~$\uparrow$ & Alpaca-Eval V2~$\uparrow$ \\
         \midrule
         \multicolumn{3}{c}{\textsc{Meta-Llama-3.2-3B}}\\
         \midrule
         \textsc{Alpaca vanilla}  & 5.52 & 6.17 / 3.54 \\
         \textsc{Alpaca + SR}  & \underline{5.65} & \underline{6.18} / \underline{4.43}\\
         \textsc{Alpaca + NILE} & \textbf{5.94} & \textbf{6.61} / \textbf{5.10}\\
         \midrule
         \textsc{Orca vanilla}  & 2.20 & 5.61 / 4.41\\
         \textsc{Orca + SR}  & \underline{3.06} & \underline{6.18} / \underline{5.51}\\
         \textsc{Orca + NILE} & \textbf{4.63} & \textbf{10.77} / \textbf{8.46}\\
         \midrule
         \multicolumn{3}{c}{\textsc{Qwen2.5-7B}}\\
         \midrule
         \textsc{Alpaca vanilla }  & \underline{7.13} & 13.84 / 7.83\\
         \textsc{Alpaca + SR}  & 6.78 & \underline{15.40} / \underline{8.90}\\
         \textsc{Alpaca + NILE}  & \textbf{8.13} & \textbf{17.42} / \textbf{12.24}\\
         \midrule 
         \textsc{Orca vanilla}  & 6.60 & \underline{19.19} / 13.42\\
         \textsc{Orca + SR} & \underline{7.05} & 18.45 / \underline{14.34}\\
         \textsc{Orca + NILE}  & \textbf{7.31} & \textbf{20.55} / \textbf{16.58}\\
         \midrule
         \multicolumn{3}{c}{\textsc{Qwen2.5-14B}}\\
         \midrule
         \textsc{Alpaca vanilla }  & 7.33 & 15.37 / 8.16\\
         \textsc{Alpaca + SR}  & \underline{7.73} & \underline{21.56} / \underline{12.59}\\
         \textsc{Alpaca + NILE}  & \textbf{8.06} & \textbf{28.55} / \textbf{17.45}\\
         \midrule 
         \textsc{Orca vanilla}  & 7.68 & 19.99 / 17.44\\
         \textsc{Orca + SR} & \underline{7.90} & \underline{24.40} / \underline{18.85}\\
         \textsc{Orca + NILE}  & \textbf{8.21} & \textbf{32.12} / \textbf{29.82}\\
        \bottomrule
    \end{tabular}
    }
    \caption{Experiment results of more LLMs on Alpaca and OpenOrca datasets. The highest values are \textbf{bolded}, and the second highest is \underline{underlined}. Complete results on more benchmarks are placed in Table~\ref{tab:more_exp_on_llm_part2}.}
    \vspace{-0.4cm}
    \label{tab:more_exp_on_llm}
\end{table}

\subsection{Ablation Study}\label{sec:ablation_st}
\begin{table}[!ht]
    \centering
    \small
    \begin{adjustbox}{max width=0.48\textwidth}
    \setlength{\tabcolsep}{3.2mm}{
    \begin{tabular}{lcccc}
    \toprule
         Method  
         &  A.-H.~$\uparrow$ & A.-E. V2~$\uparrow$ & MTB.~$\uparrow$ & BBH~$\uparrow$ \\
         \midrule
         \textsc{A.+K. w. FD} & \textbf{4.80} & 10.75 / 9.38 & 6.67 & \underline{60.73} \\
          \textsc{A.+K. w. FS 1 IKE}  & \underline{4.50} & \textbf{11.20} / \underline{9.75} & \underline{6.72} & 59.25 \\
         \textsc{A.+K. w. FS 2 IKE}  & \underline{4.50} & \underline{10.82} / \textbf{10.56} & \textbf{6.76} & \textbf{61.40} \\
         \midrule 
         \textsc{O.+K. w. FD} & \underline{5.20} & \textbf{13.67} / \underline{11.21} & \textbf{7.51} & \underline{64.03}\\
         \textsc{O.+K. w. FS 1 IKE}  & 4.90 & 12.46 / 10.99 & 7.40 & 63.89 \\
         \textsc{O.+K. w. FS 2 IKE} & \textbf{5.50} & \underline{13.00} / \textbf{11.50} & \underline{7.43} & \textbf{64.29} \\
        \bottomrule

    \end{tabular}
    }
    \end{adjustbox}
    \caption{Effects of IKE with different fewshot numbers~(\textsc{FS}) in \textsc{Llama-3}. The highest values are \textbf{bolded}, and the second highest is \underline{underlined}. For brevity, \textsc{Alpaca + KSR} and \textsc{Orca + KSR} are abbreviated as \textsc{A.+K.} and \textsc{O.+K.}, respectively.}
    \label{tab:ablation_fewshot}
\end{table}
\begin{table}[ht!]
    \centering
    \small
    \begin{adjustbox}{max width=0.48\textwidth}
    \setlength{\tabcolsep}{2.8mm}{
    \begin{tabular}{lcccc}
    \toprule
         Method  
         &  A.-H.~$\uparrow$ & A.-E. V2~$\uparrow$ & MTB.~$\uparrow$ & BBH~$\uparrow$ \\
         \midrule
         \textsc{A.+N. wo. ICF}  & \underline{4.50} & \textbf{10.82} / \textbf{10.56} & 6.76 & 61.40 \\ 
         \textsc{A.+N. w. ICF (low)}  & \textbf{4.80} & \underline{10.69} / \underline{10.43} & \textbf{6.90} & 61.40 \\
         \textsc{A.+N. w. ICF (med.)}  & 4.30 & \underline{9.99} / 9.81 & 6.65 & \underline{61.56} \\
         \textsc{A.+N. w. ICF (high)}  & \underline{4.50} & 9.92 / 9.70 & \underline{6.79} & \textbf{61.71} \\
         \midrule 
        
        \textsc{O.+N. wo. ICF}  & \underline{5.50} & 13.00 / 11.50 & \underline{7.43} & \textbf{64.29} \\
        \textsc{O.+N. w. ICF (low)}  & \textbf{6.00} & \textbf{13.70} / \textbf{12.11} & \textbf{7.48} & 64.05 \\
        \textsc{O.+N. w. ICF (med.)}  & 5.00 & \underline{13.27} / \underline{11.59} & \underline{7.43} & \underline{64.09} \\
        \textsc{O.+N. w. ICF (high)}  & 4.80 & 13.19 / 11.49 & 7.30 & 63.95 \\
        \bottomrule
    \end{tabular}
    }
    \end{adjustbox}
    \caption{Effects of ICF using different $\beta$ parameters in \textsc{Llama-3}. The highest values are \textbf{bolded}, and the second highest is \underline{underlined}. \textsc{Alpaca + NILE} and \textsc{Orca + NILE} are abbreviated as \textsc{A.+N.} and \textsc{O.+N.} for simplicity.}
    \vspace{-0.5cm}
    \label{tab:filtering_ablation_exp}
\end{table}
\paragraph{Effects of Different Internal Knowledge Sources}
We closely examine the effect of introducing LLMs' internal knowledge into \textsc{NILE} by switching the original internal knowledge source from \textsc{Mistral} to that from \textsc{Llama-3} in KSR~(extracted by \textsc{Fixed Demonstration~(FD)} prompting described in Appendix~\ref{sec:fd}). Table~\ref{tab:ablation_ik_extract} shows the comprehensive advantage of using \textsc{Llama-3}'s internal knowledge over using \textsc{Mistral}'s. Switching from \textsc{Mistral} to \textsc{Llama-3} increases \textbf{Arena-Hard} by 1.6 and 1.2 points in \textsc{Llama-3} model on the Alpaca and Orca dataset. It is also interesting to see that using internal knowledge from \textsc{Mistral} has a huge negative impact on \textsc{Llama-3} on the \textbf{BBH} task
requiring expert knowledge and complex reasoning, further highlighting the importance of such consistency. This suggests that maintaining general consistency between world knowledge from datasets and LLM internal knowledge is of necessity in effective IFT. 
\begin{table}[!h]
    \centering
    \small
    \begin{adjustbox}{max width=0.48\textwidth}
    \setlength{\tabcolsep}{3.2mm}{
    \begin{tabular}{lcccc}
    \toprule
         Method  
         &  A.-H.~$\uparrow$ & A.-E. V2~$\uparrow$ & MTB.~$\uparrow$ & BBH~$\uparrow$ \\
         \midrule
         \textsc{A.+K.~(Mistral)}  & 4.00 &  9.14 / 7.29 & 6.64 & 57.67 \\
         \textsc{A.+K.~(Llama)}  & \textbf{4.80} & \textbf{10.75} / \textbf{9.38} & \textbf{6.67} & \textbf{60.73} \\
         \midrule 
         \textsc{O.+K.~(Mistral)} & 5.10 & 12.50 / 10.25 & 5.93 & 22.32 \\
         \textsc{O.+K.~(Llama)} & \textbf{5.20} & \textbf{13.67} / \textbf{11.21} & \textbf{7.51} & \textbf{64.03} \\
        \bottomrule

    \end{tabular}
    }
    \end{adjustbox}
    \caption{Effects of KSR in \textsc{Llama-3} finetuned with internal knowledge from different LLMs. The highest values are \textbf{bolded}. Here \textsc{Alpaca + KSR} and \textsc{Orca + KSR} are abbreviated as \textsc{A.+K.} and \textsc{O.+K.} for brevity.}
    \vspace{-0.3cm}
    \label{tab:ablation_ik_extract}
\end{table}
\paragraph{Effects of IKE Fewshot Number}
Table~\ref{tab:ablation_fewshot} examines how different few-shot numbers of demonstration learning in IKE affect LLM performance. Here we evaluate three variants: 1)~\textsc{w. FD}, which extracts LLM's internal knowledge with a fixed 2-shot demonstration described in~\ref{sec:fd}; 2)~\textsc{w. FS 1 IKE}, which retrieves the top 1 most similar samples with BM25 as demonstrations; and 3)~\textsc{w. FS 2 IKE}, which retrieves the top 2 most similar samples with BM25 as demonstrations; Though \textsc{w. FS 2 IKE} leads to degradation in some benchmarks, such as \textbf{Arena-Hard} for \textsc{Alpaca} and \textbf{Alpaca-Eval} for \textsc{Orca}, it still achieves overall improvements with \textbf{BBH} for \textsc{Alpaca} increasing by 0.7 and \textbf{Arena-Hard} for \textsc{Orca} increasing by 0.3.  The results show that IKE is necessary for unaligned LLMs to more effectively extract internal knowledge, while fixed prompting reaches subpar performance. 
\paragraph{Effects of KSR}
\begin{figure}[!ht]
    \centering
    \includegraphics[width=0.42\textwidth]{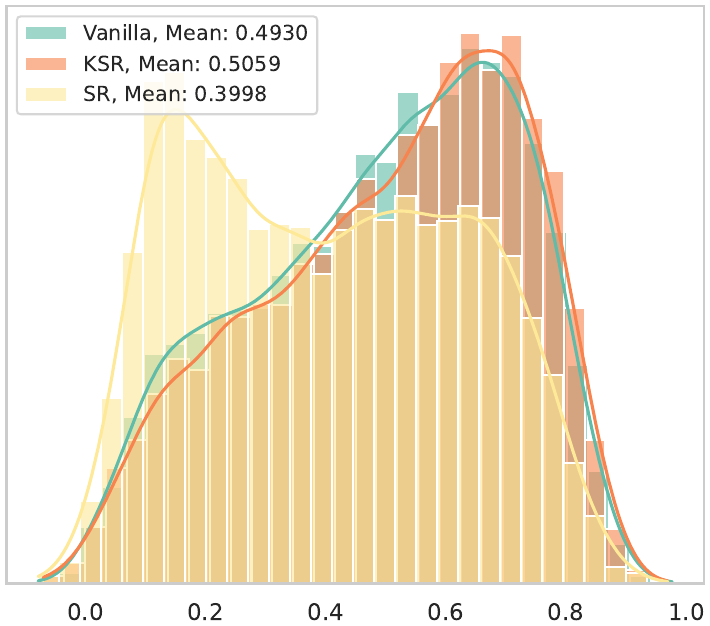}
    \caption{Distribution plot of sentence embedding similarity score in \textsc{Alpaca} dataset for \textsc{Mistral} model.}
    \label{fig:kar}
    \vspace{-0.3cm}
\end{figure}
We evaluated KSR's effectiveness in improving internal consistency between world knowledge from instructions and the model's internal knowledge. Our experiments assessed the degree to which responses incorporated internal knowledge across various models. We compared the models' vanilla, \textsc{KSR}-generated, and \textsc{SR}-generated responses for 10K randomly sampled instructions by calculating sentence similarity scores.
As shown in Figure~\ref{fig:kar}, outputs generated by \textsc{KSR} exhibit a similarity score distribution significantly closer to 1 compared to \textsc{SR} and the vanilla baseline, with Chi-squared test p-values below 0.01. Additional results in \ref{sec:effect_kar} further validate these findings. These results strongly support the effectiveness of \textsc{KSR} in enhancing internal consistency by integrating world and internal knowledge.

\paragraph{Effects of ICF}
Table~\ref{tab:filtering_ablation_exp} looks into the effect of ICF. $\beta$ is set to 1-st percentile in \textsc{Alpaca + NILE w. ICF (low)} and to 2-nd percentile in \textsc{Orca + NILE w. ICF (low)}. We set $\beta$ to 5-th and 10-th percentile for \textsc{NILE w. ICF~(medium)} and \textsc{NILE w. ICF~(high)}. The results empirically prove that striking a balance between consistent and inconsistent knowledge in the IFT dataset is necessary for \textsc{NILE} to achieve ideal performance. We find the general advantage of \textsc{ALPACA + NILE w. ICF (low)} over \textsc{ALPACA + NILE w. ICF (medium)} and \textsc{ALPACA + NILE w. ICF (high)} and discarding ICF (\textsc{ALPACA + NILE wo. ICF}), indicating that a surplus of overly consistent or inconsistent samples in IFT datasets both hurt LLM's performance, and it is crucial to find the middle ground in these samples. This experiment further verifies our design choices of the ICF phase.

\section{Conclusion}
We present \textsc{NILE}, an innovative framework designed to enhance the training efficacy of IFT datasets by aligning them with LLMs' internal knowledge. Our extensive experiments demonstrate substantial improvements across various benchmarks, highlighting the crucial role of maintaining consistency between models' internal knowledge and external knowledge in datasets. Each component of the \textsc{NILE} framework has been validated, reinforcing its importance in achieving better alignment. \textsc{NILE} offers promising directions for boosting the capabilities of LLMs and unlocking their full potential.

\section*{Limitations}
While NILE can already obtain satisfactory performance, future works should expand NILE's training by utilizing the complete OpenOrca dataset rather than the current 50,000-sample subset (5\% of the dataset), due to limited time and computational resources. To ensure a fair comparison of experiments in our study, we maintain a consistent dataset size by reverting to the original answer rather than discarding samples during the ICF phase. In future work, we aim to explore more advanced data selection techniques for the ICF process. Additionally, future research should examine NILE's capability for iterative instruction refinement, as the current implementation uses only a single revision pass. These expansions could further enhance NILE's instruction-following capabilities.

\section*{Ethics Statement}
We conducted this study strictly under the guidance of community ethical principles. 
The utilized IFT datasets are reported to be safe and free of content that may contain discrimination, personally identifiable information, or any other undesirable behaviors.
We meticulously curate our instructions to the LLMs to ensure that the tasks are limited to knowledge generation and knowledge-relevant revisions,  thereby avoiding content that may pose ethical concerns.

\section*{Acknowledgement}
The work described in this paper was partially supported by the Research Grants Council of the Hong Kong Special Administrative Region, China (CUHK 2410072, RGC R1015-23). As the first author, I would like to express my heartfelt gratitude to my family, co-authors, and advisor, Prof. Irwin King, for their unwavering support and invaluable guidance throughout this work.

I would also like to dedicate this acknowledgment to the memory of \begin{CJK*}{UTF8}{bsmi}\textit{方大同~(Khalil Fong)},
\end{CJK*}
an extraordinarily talented and influential R\&B singer-songwriter. During the countless days and nights spent on this research and my PhD journey, his music and life philosophy provided me with comfort and strength to overcome every obstacle. His passing on February 21, 2025, was a profound loss for Chinese pop music and for me personally. I believe his vision and influence will continue to accompany us, transcending time and mortality, into the distant future. 
\begin{CJK*}{UTF8}{bsmi}\textit{成長是永遠，離別是空懸。在千尋之外，我依然存在。~(Growth is everlasting, and parting is always in suspense. Beyond searches, I still exist.)}
\end{CJK*}
\bibliography{anthology_nile_rebibed,custom_nile_rebibed}

\appendix

\section{Appendix}
\label{sec:appendix}

\subsection{Implementation Details of \textsc{NILE}}\label{sec:imple_nile}
For all experiments in this work, we use the Python 3.10.14 environment and vLLM 0.5.5 library\footnote{\url{https://github.com/vllm-project/vllm}} for LLM local inference of both \textsc{Mistral} and \textsc{Llama-3}. For vLLM inference hypermeters, we set the random seed to 42, max\_tokens to 1024, temperature to 0.7, top\_k to 50, top\_p to 0.7, and repetition\_penalty to 1. We run all experiments on a server with an Intel Xeon Silver 4309Y CPU and 8 Nvidia RTX A6000 GPU having 48GB GDDR6 VRAM, and we utilize official checkpoints \textsc{Mistral-7B-v0.3}\footnote{\url{https://huggingface.co/mistralai/Mistral-7B-v0.3}} for \textsc{Mistral} and \textsc{Meta-Llama-3.1-8B}\footnote{\url{https://huggingface.co/meta-llama/Llama-3.1-8B}} for \textsc{Llama-3}.
For LLM instruction fine-tuning in this work, we choose llama-recipes\footnote{\url{https://github.com/meta-llama/llama-recipes}} for \textsc{Llama-3} and alignment-handbook\footnote{\url{https://github.com/huggingface/alignment-handbook}} for \textsc{Mistral}. For \textsc{Llama-3} fine-tuning, we set context\_length to 2048, gradient\_accumulation\_step to 32, learning rate to 2e-5, and training batch size to 4. As for \textsc{Mistral} fine-tuning, we set context\_length to 2048, gradient\_accumulation\_step to 32, learning rate to 2e-5, training batch size to 4, lr\_scheduler\_type to "cosine", num\_train\_epochs to 3, and warmup\_ratio to 0.1. Fine-tuning for both \textsc{Llama-3} and \textsc{Mistral} is done within 5 hours using 8 A6000 GPU.
\subsubsection{Internal Knowledge Extraction (IKE)}
For demonstration sample, we randomly sample $m = 5,000$ instruction pairs $q^d_i =\{\mathrm{instruction}^{d}_i, \mathrm{input}^{d}_i\}$ from Alpaca dataset\footnote{\url{https://huggingface.co/datasets/vicgalle/alpaca-gpt4}}, since instructions in it are simple and straightforward, which is suitable for LLM demonstration learning. We leverage \textsc{gpt-4-turbo-2024-04-09} through the OpenAI API for generating demonstrations given $q^d_i$ shown in Table~\ref{tab:demo_gen_prompt}. For GPT-4 endpoints, we use the OpenAI 1.42.0 Python library and set n to 1, temperature to 0.7, and max\_tokens to 1,024. We stick to the regulations from the OpenAI company when accessing its API. For retriever $\mathcal{R}$ in IKE, we choose the BM25 and Contriver implementation\footnote{\url{https://github.com/castorini/pyserini}} from the Pyserini 0.38.0 library. 
\subsubsection{Design Choice: BM25 vs Contriver}
\label{sec:bm25_vs_contriver}
\begin{table*}[!ht]
    \centering
    \setlength{\tabcolsep}{3.2mm}{
    \begin{tabular}{lcccc}
    \toprule
         Method  
         &  Arena-Hard~$\uparrow$ & Alpaca-Eval V2~$\uparrow$ & MTBench~$\uparrow$ & BBH~$\uparrow$ \\
         \midrule
        \textsc{Orca + NILE w. Contriver}  & 4.70 & \textbf{14.63} / \textbf{12.17} & 7.29 & 64.00 \\
        \textsc{Orca + NILE w. BM25}  & \textbf{5.50} & 13.00 / 11.50 & \textbf{7.43} & \textbf{64.29} \\
        \bottomrule
    \end{tabular}
    }
    \caption{Comparison between choosing BM25 and Contriver. The highest values are \textbf{bolded}.}
    \label{tab:bm25_vs_contriver}
\end{table*}
The performance gain of the NILE choosing BM25 over Contriver in IKE is shown in Table~\ref{tab:bm25_vs_contriver}.
\subsubsection{IKE Sample Demonstration}
\label{sec:ike_sample_demon}
Table~\ref{tab:sample_demo} illustrates the sample 2-shot demonstration set from IKE, and Table~\ref{tab:ik_db_demo} shows two samples from the demonstration database $\mathcal{F}^{demo}$ in IKE.
\begin{table*}[!ht]
    \centering
    \begin{tcolorbox}[colframe=black, colback=gray!10!white, coltitle=black, boxrule=0.5mm]
    \textbf{Instruction $\mathrm{instruction}^o$:} \textit{Recommend 3 books that could get someone into a new field of study.}\\
    \textbf{Input $\mathrm{input}^o$:} None\\
    \textbf{2-shot Demonstration $ f^{demo}(q^o)=\{(q^{\mathcal{R}}_{1}, ik^{\mathcal{R}}_{1}), (q^{\mathcal{R}}_{2}, ik^{\mathcal{R}}_{2})\}$:}\\
    \tcbline
    $\mathbf{q^{\mathcal{R}}_1}:$ Provide 3 pieces of advice for someone who is struggling to stay motivated in their studies.\\
    $\mathbf{ik^{\mathcal{R}}_1}:$ 1. Understanding Motivation: It is essential to comprehend the concept of motivation, including intrinsic and extrinsic motivation. Intrinsic motivation refers to doing something because it is inherently interesting or enjoyable, while extrinsic motivation involves doing something because it leads to a separable outcome. Understanding these types of motivation can help identify what drives an individual's desire to study and how to bolster it.\\2. Goal Setting: Knowledge of effective goal-setting strategies is crucial. Setting specific, measurable, achievable, relevant, and time-bound (SMART) goals can help create a clear roadmap for studying. This process can help break down the overwhelming feeling into manageable tasks, making it easier to stay motivated.\\
    ......\\
    \tcbline
    $\mathbf{q^{\mathcal{R}}_2}:$ Please recommend a movie for someone who likes animated films.\\
    $\mathbf{ik^{\mathcal{R}}_2}:$ 1. Understanding Animated Films: Animated films, also known as cartoons, use drawn or computer-generated imagery to create the illusion of movement. They can be in 2D, 3D, stop-motion or other animation techniques. Knowledge of the different styles and techniques of animation is crucial to recommending films that align with specific preferences.\\2. Popular Animation Studios: Some of the most prominent animation studios include Disney, Pixar, DreamWorks, Studio Ghibli, and Illumination. Each studio has its own unique style and storytelling approach. Familiarity with the filmographies of these studios can help in making informed recommendations.\\3. Genres within Animation: Animated films encompass a wide range of genres, from adventure and fantasy to drama and comedy. Some animated films are targeted towards children, while others might appeal to a broader age range including adults. Understanding the various genres and their target audiences can guide the recommendation process.\\
    4. Cultural and Thematic Elements: Animated films often incorporate diverse cultural stories and themes. Films might explore complex themes like identity, family, and morality, or might be more light-hearted and comedic. Awareness of these elements can help match a film to a viewer's personal interests and values.\\
    ......
    \end{tcolorbox}
    \caption{Sample 2-shot demonstration in IKE.}
    \label{tab:sample_demo}
\end{table*}
\begin{table*}[!ht]
    \centering
    \begin{tcolorbox}[colframe=black, colback=gray!10!white, coltitle=black, boxrule=0.5mm]
    $\mathbf{q^d_1}:$ What would be the best type of exercise for a person who has arthritis?\\
    $\mathbf{ik^d_1}:$ 1. Understanding Arthritis: Arthritis is a chronic condition characterized by inflammation in the joints, which can cause pain, stiffness, and reduced mobility. There are different types of arthritis, such as osteoarthritis and rheumatoid arthritis, each with varying symptoms and treatment approaches.\\2. Impact of Exercise on Arthritis: Exercise is generally considered beneficial for individuals with arthritis. It can help to reduce joint pain, increase flexibility, improve muscle strength, enhance endurance, and promote overall physical function.\\3. Types of Exercise Suitable for Arthritis:\\   - Low-impact Aerobic Activities: These exercises are gentle on the joints and include walking, swimming, and cycling. They help in cardiovascular conditioning without putting excessive stress on the joints.\\   - Resistance Training: Using light weights or resistance bands can help strengthen the muscles around the joints, providing better support and reducing the burden on the joints.\\   - Flexibility Exercises: Activities such as stretching and yoga can improve joint flexibility and range of motion, helping to alleviate stiffness.\\
    ......\\
    \tcbline
    $\mathbf{q^d_2}:$ Calculate the atomic mass for lithium. \\
    $\mathbf{ik^d_2}:$ 1. Understanding Atomic Mass: Atomic mass, also known as atomic weight, is defined as the weighted average mass of atoms of an element based on the abundance of each isotope of the element in nature. It is usually measured in atomic mass units (amu).\\2. Isotopes: Isotopes are atoms of the same element that have the same number of protons but different numbers of neutrons. This results in different mass numbers for each isotope. The atomic mass of an element is calculated by taking into account the masses and relative abundances of all its naturally occurring isotopes.\\3. Lithium Isotopes: Lithium has two stable isotopes, lithium-6 (6Li) and lithium-7 (7Li). These isotopes differ in their neutron count, affecting their individual atomic masses. Lithium-6 has 3 neutrons, while lithium-7 has 4 neutrons.\\4. Natural Abundance: The natural abundance of an isotope refers to the percentage of that isotope found naturally in a sample of the element. For lithium, lithium-7 is more abundant than lithium-6. The exact percentages of natural abundance can vary slightly depending on the source, but generally, lithium-7 accounts for about 92.5\% while lithium-6 is about 7.5\%.\\
    ......
    \end{tcolorbox}
    \caption{Samples from demonstration database $\mathcal{F}^{demo}$ in IKE.}
    \label{tab:ik_db_demo}
\end{table*}
\subsubsection{Knowledge-aware Sample Revision (KSR)}
For KSR, we also use \textsc{gpt-4-turbo-2024-04-09} endpoint as revisor agent $\mathcal{A}_r$. We use the OpenAI 1.42.0 Python library with n set to 1, temperature to 0.7, and max\_tokens to 1,024. We run KSR on 52,000 samples from the Alpaca dataset and 50,000 samples from the OpenOrca dataset. Case studies of KSR can be found in Table~\ref{tab:case_study_ksr}. These results display the KSR's capability to infuse internal knowledge information into original answers through revision.
\subsubsection{Sample Revision (SR)}
\label{sec:sr}
Unlike KSR, \textsc{Sample Revision (SR)} does not revise for each instruction pair $(q^o, a^o)$. Therefore, $\mathcal{A}_r$ in SR only uses external knowledge such as world knowledge from $(q^o, a^o)$ and its own parameter knowledge, being completely isolated from internal knowledge $ik$ of $\mathcal{M}$. Table~\ref{tab:sr_prompt} shows the detailed prompt of the revisor $\mathcal{A}_r$ in SR.
\begin{table}[!ht]
    \centering
    \begin{tcolorbox}[colframe=black, colback=gray!10!white, coltitle=black, boxrule=0.5mm]
    Provide a better response based on "$\mathbf{\{a^o\}}$" to comply with given instruction, input, and related knowledge.\\\\
    Instruction: $\mathbf{\{\mathrm{\textbf{instruction}}^{o}\}}$\\
    Input:$\mathbf{\{ \mathrm{\textbf{input}}^{o}\}}$\\
    \\
    Please directly output the improved response.
    \end{tcolorbox}
    \caption{Prompt for Sample Revision.}
    \vspace{-0.5cm}
    \label{tab:sr_prompt}
\end{table}
\subsubsection{Statistics of Inference Overhead}
\label{sec:infer_o}
In the IKE step of NILE, internal knowledge is efficiently extracted from sample datasets within 6 hours using vLLM on an 8-A6000 GPU server. To further show how much inference overhead is introduced, we
measured the average token usage per sample for this step, which is detailed in Table~\ref{tab:ike_token_num}. 
\begin{table}[!ht]
    \centering
    \setlength{\tabcolsep}{2mm}{
    \begin{tabular}{l|cc}
    \toprule
         Model / Dataset  
         & \textsc{Llama} & \textsc{Mistral} \\
         \midrule
         \textsc{OpenOrca} & 684.8 & 801.1\\
         \textsc{Alpaca}  & 620.6 & 544.5\\
    \bottomrule

    \end{tabular}
    }
    \caption{Average generated tokens per sample using vLLM during IKE across different datasets and models.}
    \label{tab:ike_token_num}
\end{table}

The Knowledge-aware Sample Revision (KSR) step further optimizes efficiency, with GPT-4 achieving modest token usage per sample, also shown in Table~\ref{tab:ksr_token_num}. This results in exceptionally low operational costs, making our approach scalable, cost-effective, and practical for real-world applications.
\begin{table}[!ht]
    \centering
    \setlength{\tabcolsep}{2mm}{
    \begin{tabular}{l|cc}
    \toprule
         Model / Dataset  
         & \textsc{Llama} & \textsc{Mistral} \\
         \midrule
         \textsc{OpenOrca} & \$0.015/182.0 & \$0.016/182.1\\
         \textsc{Alpaca}  & \$0.013/193.7 & \$0.011/165.3\\
    \bottomrule

    \end{tabular}
    }
    \caption{Price (in USD) and average generated tokens per sample during KSR across different datasets and models.}
    \label{tab:ksr_token_num}
\end{table}


\subsection{Evaluating NILE's Effectiveness on External Knowledge-Intensive Tasks}
To briefly examine some of NILE's potential issues and limitations, we conducted an additional experiment on the SQuADv2~\cite{rajpurkar2018know} validation set using our sampled Alpaca-GPT4 dataset under the same settings outlined in our paper. The SQuADv2 validation set was chosen because it contains 119,000 test samples of reading comprehension, where large language models (LLMs) must answer questions based on external knowledge provided in corresponding supporting passages. As such, it serves as a suitable and rigorous benchmark for evaluating an LLM's ability to comprehend and utilize external knowledge effectively. The results in Table~\ref{tab:squadv2_res} demonstrate that NILE can positively influence this capability.

\subsection{Examining NILE's Performance on Multitask Accuracy}
In order to further test the extensiveness of NILE's improvement on LLMs,  we have conducted additional experiments on MMLU using two models (\textsc{Meta-LLama-3.1-8B} and \textsc{Mistral-7B-v0.3}) and two IFT datasets (Alpaca and Orca). All experiments adhered to the official default configuration from the lm-evaluation-harness implementation\footnote{\url{https://github.com/EleutherAI/lm-evaluation-harness/tree/main/lm_eval/tasks/mmlu}}. The results presented in Table~\ref{tab:mmlu_res} demonstrate that NILE consistently achieves significant performance improvements in highly complex QA settings like MMLU. Interestingly, we observed a noticeable dip in accuracy with the SR baseline across both datasets and models. This result further underscores the necessity of incorporating internal knowledge within the NILE framework to enhance alignment in IFT datasets.
\begin{table}[!ht]
    \centering
    \setlength{\tabcolsep}{2mm}{
    \begin{tabular}{lc}
    \toprule
         Method  
        & Accuracy~(\%)~$\uparrow$ \\
         \midrule
         \multicolumn{2}{c}{\textsc{Mistral-7b-v0.3}}\\
         \midrule
         \textsc{Alpaca Vanilla}  & 57.21\\
         \textsc{Alpaca + SR}  & 56.25\\
         \textsc{Alpaca + NILE}  & \textbf{57.56}\\
         \midrule
         \textsc{Orca Vanilla}  & \textbf{56.92}\\
         \textsc{Orca + SR}  & 54.54\\
         \textsc{Orca + NILE}  & 56.91\\
         \midrule
         \multicolumn{2}{c}{\textsc{Meta-Llama-3.1-8B}}\\
         \midrule
         \textsc{Alpaca Vanilla}  & 62.51\\
         \textsc{Alpaca + SR} & 62.41\\
         \textsc{Alpaca + NILE} & \textbf{63.93}\\
         \midrule
         \textsc{Orca Vanilla}  & 62.68\\
         \textsc{Orca + SR} & 62.19\\
         \textsc{Orca + NILE} & \textbf{63.09}\\
        \bottomrule
    \end{tabular}
    }
    \caption{Experiment results of NILE on MMLU benchmark. The highest values are \textbf{bolded}.}
    \label{tab:mmlu_res}
\end{table}

\begin{table}[!ht]
    \centering
    \setlength{\tabcolsep}{1.6mm}{
    \begin{tabular}{lcc}
    \toprule
         Method  
        & EM~$\uparrow$ & F1~$\uparrow$ \\
         \midrule
         \multicolumn{3}{c}{\textsc{Mistral-7b-v0.3}}\\
         \midrule
         \textsc{Alpaca vanilla }  & 4.91 & 14.13\\
         \textsc{Alpaca + NILE}  & \textbf{5.61} & \textbf{14.51}\\
         \midrule
         \multicolumn{3}{c}{\textsc{Meta-Llama-3.1-8B}}\\
         \midrule
         \textsc{Alpaca vanilla}  & 1.41 & 9.35 \\
         \textsc{Alpaca + NILE} & \textbf{4.51} & \textbf{11.65}\\
        \bottomrule
    \end{tabular}
    }
    \caption{Experiment results of NILE on SQuADv2 dataset. The highest values are \textbf{bolded}.}
    \vspace{-0.5cm}
    \label{tab:squadv2_res}
\end{table}
\begin{table*}[!ht]
    \centering
    \setlength{\tabcolsep}{1.6mm}{
    \begin{tabular}{lcccc}
    \toprule
         Method  
        & Arena-Hard~$\uparrow$ & Alpaca-Eval V2~$\uparrow$ & MTBench~$\uparrow$ & BBH~$\uparrow$ \\
         \midrule
         \multicolumn{5}{c}{\textsc{Meta-Llama-3.2-3B}}\\
         \midrule
         \textsc{Alpaca vanilla} & 3.50 & 6.17 / 3.54 & 5.52 & \underline{43.63} \\
         \textsc{Alpaca + SR}  & \underline{3.00}  & \underline{6.18} / \underline{4.43} & \underline{5.65} & 43.02\\
         \textsc{Alpaca + NILE} & \textbf{4.60} & \textbf{6.61 / 5.10} & \textbf{5.94} & \textbf{43.74}\\
         \midrule
         \textsc{Orca vanilla}  & 3.60 & 5.61 / 4.41 & 2.20 & \underline{43.36}\\
         \textsc{Orca + SR}  & \underline{4.00} & \underline{6.18} / \underline{5.51} & \underline{3.06} & 41.87\\
         \textsc{Orca + NILE} & \textbf{4.20} & \textbf{10.77 / 8.46} & \textbf{4.63} & \textbf{45.69}\\
         \midrule
         \multicolumn{5}{c}{\textsc{Qwen2.5-7B}}\\
         \midrule
         \textsc{Alpaca vanilla }  & 11.90 & 13.84 / 7.83 & \underline{7.13} & \textbf{45.95}\\
         \textsc{Alpaca + SR}  & \underline{14.40} & 
         \underline{15.40} / \underline{8.90} & 6.78 &  \underline{45.94}\\
         \textsc{Alpaca + NILE}  & \textbf{18.40} & \textbf{17.42 / 12.24} & \textbf{8.13} & 45.72\\
         \midrule 
         \textsc{Orca vanilla}  & \underline{14.90} & \underline{19.19} / 13.42 & 6.60 & 46.89\\
         \textsc{Orca + SR} & 13.60 & 18.45 / \underline{14.34} & \underline{7.05} & \textbf{49.32}\\
         \textsc{Orca + NILE}  & \textbf{17.10} & \textbf{20.55 / 16.58} & \textbf{7.31} & \underline{48.61}\\
         \midrule
         \multicolumn{5}{c}{\textsc{Qwen2.5-14B}}\\
         \midrule
         \textsc{Alpaca vanilla }  & 12.80 & 15.37 / 8.16 & 7.33 & 48.01\\
         \textsc{Alpaca + SR}  & \underline{20.00} & \underline{21.56} / \underline{12.59} & \underline{7.73} & \underline{48.69}\\
         \textsc{Alpaca + NILE}  & \textbf{22.60} & \textbf{28.55 / 17.45} & \textbf{8.06} & \textbf{49.01}\\
         \midrule 
         \textsc{Orca vanilla}  & 24.00 & 19.99 / 17.44 & 7.68 & 48.49\\
         \textsc{Orca + SR} & \underline{26.60} & \underline{24.40} / \underline{18.85} & \underline{7.90} & \underline{47.66}\\
         \textsc{Orca + NILE}  & \textbf{31.90} & \textbf{32.12 / 29.82} & \textbf{8.21} & \textbf{49.96}\\
        \bottomrule
    \end{tabular}
    }
    \caption{Complete experiment results of more LLMs on Alpaca and OpenOrca datasets for \textbf{Arena-Hard}, \textbf{Alpaca-Eval V2 LCWR}, \textbf{MTBench}, and \textbf{BBH} benchmarks. The highest values are \textbf{bolded}, and the second highest is \underline{underlined}.}
    \vspace{-0.3cm}
    \label{tab:more_exp_on_llm_part2}
\end{table*}
\subsection{Experiment Details}
\subsubsection{Benchmarks}
We use the officially recommended settings from all benchmarks for evaluation. For Alpaca-Eval V2, we use "alpaca\_eval\_cot\_gpt4\_turbo\_fn" as annotators, and we set max\_new\_tokens to 1024, temperature to 1.0, top\_p to 1.0, and batch\_size to 128. 
For Arena-Hard, we set the temperature to 0.0, max\_tokens to 1024, judge\_model to gpt-4-1106-preview, baseline\_model to gpt-4-0314, and num\_choices to 1. For the Arena-Hard and Alpaca-Eval V2 benchmark, we keep the same alpaca-style system prompt as the fine-tuning stage during evaluation. As for BBH and MTBench, we use the default settings in the official source code.
\begin{figure*}[!ht]
    \centering
    \includegraphics[width=0.8\textwidth]{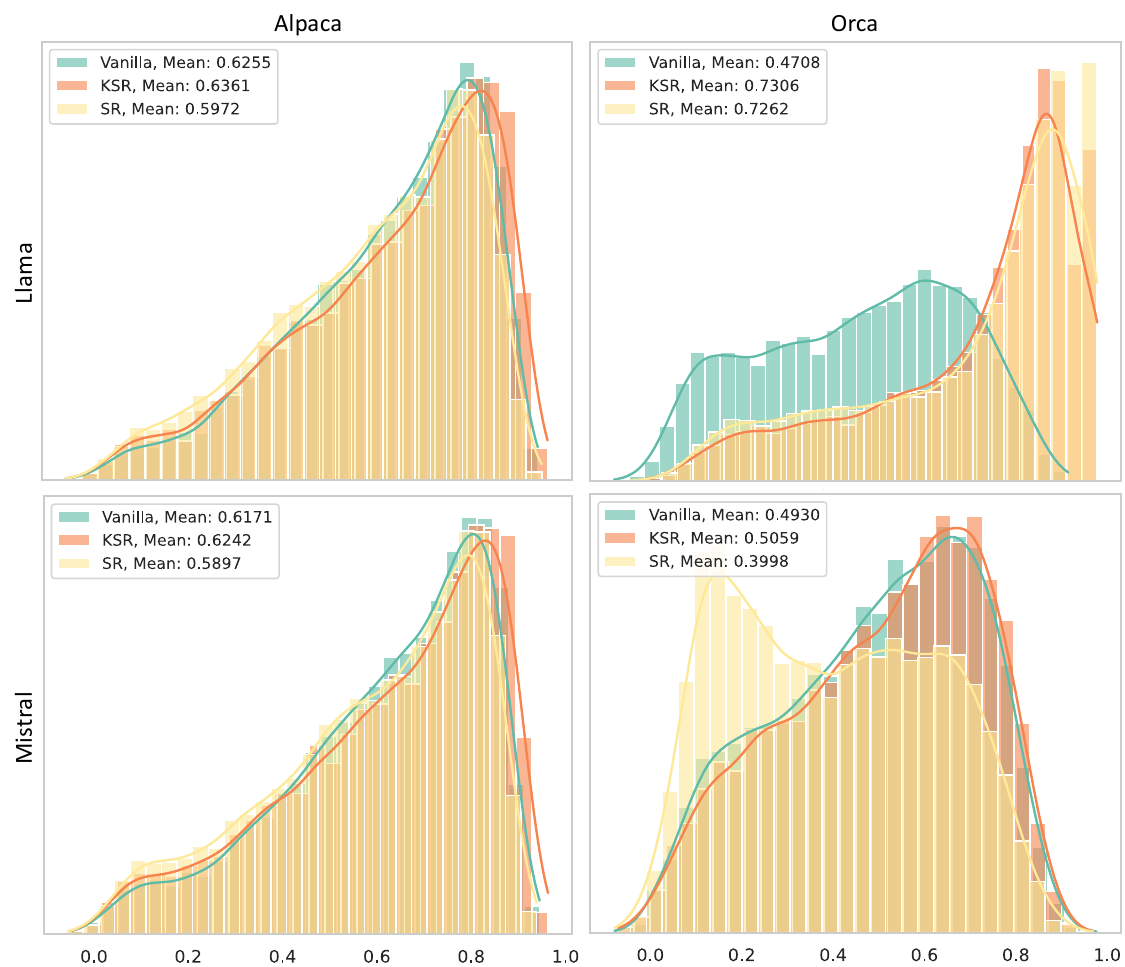}
    \caption{Distribution of sentence embedding similarity across different LLMs and IFT datasets.}
    \label{fig:kar_all}
\end{figure*}

\subsubsection{Fixed Demonstration (FD)}
\label{sec:fd}
\begin{table*}[!ht]
    \centering
    \begin{tcolorbox}[colframe=black, colback=gray!10!white, coltitle=black, boxrule=0.5mm]
    Instruction:\\
"Give three tips for staying healthy."\\
\\
Related Knowledge:\\
* Importance of health: Maintaining good health is crucial for overall well-being and quality of life.
* Factors affecting health: A person's health can be influenced by various factors such as diet, exercise, sleep, stress, and genetics.
* Prevention is key: Preventing illnesses and maintaining good health requires adopting healthy habits and making lifestyle changes.
* Healthy habits: Developing healthy habits such as regular exercise, balanced diet, and adequate sleep can help prevent chronic diseases.
* Lifestyle changes: Making lifestyle changes such as quitting smoking, reducing sugar intake, and managing stress can also contribute to good health.
* Importance of self-care: Taking care of one's physical, emotional, and mental health is essential for overall well-being.
* Access to healthcare: Having access to quality healthcare and medical facilities is also important for staying healthy.
* Healthy behaviors: Engaging in healthy behaviors such as regular check-ups, vaccinations, and screenings can help prevent illnesses and detect health issues early.
* Health literacy: Having knowledge and understanding of health-related information is important for making informed decisions about one's health.
\\
Instruction:\\
"What are the three primary colors?"
\\\\
Related Knowledge:\\
* Primary colors are colors that cannot be created by mixing other colors together.
* The three primary colors are:
        1. Red
        2. Blue
        3. Yellow
* Primary colors are the base colors used to create all other colors.
* By mixing different combinations of primary colors, you can create secondary colors, tertiary colors, and a wide range of shades and hues.
* Primary colors are often used in art, design, and painting to create bold and vibrant colors.
* The primary colors are also used in color theory to understand how colors interact with each other and how they can be used to create contrast, harmony, and balance.
* In addition to art and design, primary colors are also used in science, technology, engineering, and mathematics (STEM) fields, such as physics, chemistry, and biology, to describe and analyze the properties of light and color.
* The primary colors are a fundamental concept in many cultures and have been used in art and design for centuries, with examples found in ancient civilizations such as Egypt, Greece, and China.\\
    \\
    Instruction:\\
    $\mathbf{\{\mathrm{\textbf{instruction}}^{o}, \mathrm{\textbf{input}}^{o}\}}$\\
    \\
    Related Knowledge:
    \end{tcolorbox}
    \caption{Prompt for Fixed Demonstration (FD).}
    \label{tab:fd_prompt}
\end{table*}
Table~\ref{tab:fd_prompt} provides the prompt of the Fixed Demonstration (FD) used for extracting LLM internal knowledge in the experiments. The FD employs a fixed set of 2-shot demonstrations, serving as a baseline for IKE without incorporating demonstration learning.
\subsubsection{Effects of KSR}
\label{sec:effect_kar}

To validate KSR's effectiveness in enhancing internal consistency between world knowledge from instructions and the model's internal knowledge, we conducted experiments measuring the similarity between extracted internal knowledge and baseline knowledge across different models and datasets. In \textsc{Llama-3} and \textsc{Mistral}, we used the instructions from the Alpaca and Orca as prompts to evaluate the models' internal knowledge. Then, we obtained the models' vanilla output for these instructions, the output adjusted using \textsc{KSR}, and the output using \textsc{SR}. We randomly sampled 10,000 instructions to calculate the sentence similarity between these three outputs and the internal knowledge. As demonstrated in Figure~\ref{fig:kar_all}, the similarity score distribution of the outputs adjusted by KSR is significantly closer to 1 compared to the other two methods, with Chi-squared test p-values lower than 0.01. 

These results provide strong evidence supporting the effectiveness of the \textsc{KSR} approach in increasing the internal consistency from instructions by integrating relevant world and internal knowledge. The case study of KSR is listed in Table~\ref{tab:case_study_ksr}.
\begin{table*}[h!]
\small
\centering
\renewcommand{\arraystretch}{1.5}
\setlength{\tabcolsep}{10pt}
\begin{tabular}{|p{2cm}|p{6cm}|p{6cm}|}
\hline
\textbf{Instruction} & \textbf{Response} & \textbf{Internal Knowledge} \\ \hline

\textit{Compose a new email subject line for a following email.} \newline 
\textbf{Input:} The email body contains a request to renew a subscription. & 
\textbf{Vanilla:} "Reminder: Renew Your Subscription Today!"
\newline
\textbf{Revised:} "\textcolor{red!70!black}{Action Required:} Time to Renew Your Subscription!"
&
Email subject line: The subject line of an email is the first thing a recipient sees and can determine whether they open the email or not. It should be clear, concise, and relevant to the email content.* Subscription renewal: A subscription renewal is a request to continue receiving a service or product after the initial subscription period has ended. It is important to provide clear instructions and information about the renewal process.* Clear instructions: Providing clear instructions in the email body can help ensure that the recipient understands the request and knows what to do next. It is important to use simple language and avoid technical jargon.* Information about renewal process: Including information about the renewal process, such as the cost, payment options, and deadlines, can help the recipient make an informed decision and take the necessary steps to renew their subscription.\textcolor{red!70!black}{* Call-to-action: Including a clear call-to-action in the email body can help ensure that the recipient knows what to do next and takes the desired action. It is important to use a strong and direct call-to-action that is easy to understand and takes the recipient to the next step.}
\\ \hline

\textit{Write a sentence containing a hyperbolic metaphor.} \newline
\textbf{Input:} None & 
\textbf{Vanilla}: He was so hungry, he could eat an entire elephant.
\newline
\textbf{Revised}: He was so ravenous, \textcolor{red!70!black}{he could devour a mountain of hamburgers as tall as Everest.}
& Hyperbole is a figure of speech that uses exaggeration for emphasis or effect.\textcolor{red!70!black}{* Hyperbolic metaphors are metaphors that use hyperbole to create a vivid and dramatic image.}* For example, "The crowd was so large that it stretched from one end of the stadium to the other."* This metaphor uses hyperbole to create a dramatic image of the crowd\'s size.* Hyperbolic metaphors can be used to create a strong emotional response or to make a point more effectively.\textcolor{red!70!black}{* They can also be used to add humor or to create a sense of exaggeration or absurdity.}\\ \hline

\textit{Create a 3-5 sentence definition for the term "data science".}
\newline
\textbf{Input:} None
& 
\textbf{Vanilla:} Data science is an interdisciplinary field that combines techniques from statistics, mathematics, computer science, and information science to extract insights and knowledge from data. This field deals with both structured and unstructured data, using algorithms, predictive models, and machine learning to make informed decisions. Data science has broad applications, from businesses and marketing to healthcare and scientific research, where it is used to answer questions, find patterns, and make strategic decisions.
\newline
\textbf{Revised:} Data science is an interdisciplinary field that uses methods from statistics, mathematics, computer science, and domain-specific knowledge to extract insights and information from both structured and unstructured data. It employs techniques such as machine learning, predictive modeling, and data mining to analyze data sets, identify patterns, and make predictions. \textcolor{red!70!black}{The ultimate goal of data science is to help organizations make informed decisions, optimize processes, and solve complex problems by leveraging vast amounts of data. This field is critical across various industries, including healthcare, finance, and technology, where data-driven insights are essential for strategic decision-making and operational improvements}.
&
Data science is an interdisciplinary field that combines techniques from statistics, computer science, and domain knowledge to extract insights from data.* It involves the use of advanced analytics and machine learning algorithms to analyze large datasets and identify patterns, trends, and correlations.* Data scientists use various tools and techniques such as data mining, predictive modeling, and natural language processing to uncover insights and make predictions about future events.\textcolor{red!70!black}{* The goal of data science is to use data to solve complex problems, make better decisions, and drive business value.* Data science has become increasingly important in recent years due to the rapid growth of data and the need for organizations to make sense of it.* Data scientists play a critical role in many industries, including healthcare, finance, retail, and technology, where they help organizations gain insights from data to improve operations, optimize processes, and make better decisions.}
\\ \hline

\end{tabular}
\caption{Case study of KSR. Related KSR revisions and internal knowledge are marked in \textcolor{red!70!black}{red}.}
\label{tab:case_study_ksr}
\end{table*}
\end{document}